\documentclass[sigconf]{acmart}

\AtBeginDocument{%
  \providecommand\BibTeX{{%
    \normalfont B\kern-0.5em{\scshape i\kern-0.25em b}\kern-0.8em\TeX}}}

\setcopyright{acmcopyright}
\copyrightyear{2022}
\acmYear{2022}
\acmDOI{10.1145/1122445.1122456}

\acmConference[CF'22]{CF'22: ACM International Conference on Computing Frontiers}{May 17--19, 2022}{Turin, Piedmont, Italy}
\acmBooktitle{CF'22: ACM International Conference on Computing Frontiers,
  May 17--19, 2022, Turin, Piedmont, Italy}
\acmPrice{15.00}
\acmISBN{978-1-4503-XXXX-X/18/06}

\usepackage{threeparttable}
\def\BibTeX{{\rm B\kern-.05em{\sc i\kern-.025em b}\kern-.08em
    T\kern-.1667em\lower.7ex\hbox{E}\kern-.125emX}}
\usepackage{multirow}
\usepackage[caption=false]{subfig}
\usepackage{tabu}
\usepackage{color, colortbl}
\newcommand{\thickhline}{%
    \noalign {\ifnum 0=`}\fi \hrule height 1pt
    \futurelet \reserved@a \@xhline
}
\usepackage[linesnumbered,ruled,noend]{algorithm2e}
\usepackage{xcolor}

\SetCommentSty{mycommfont}    
\usepackage{pifont}
\usepackage{multirow}
\usepackage{hyperref}
\newcommand{\nm}{\text{{neuromorphic}}}{}
\newcommand{\ftn}{\text{{fault-tolerant}}}{}
\newcommand{\ftc}{\text{{fault tolerance}}}{}
\newcommand{\ckt}{\text{{circuitry}}}{}
\newcommand{\ckts}{\text{{circuitries}}}{}
\newcommand{\astro}{\text{{astrocyte}}}{}
\newcommand{\mubrain}{$\mu$\text{{Brain}}}{}

\usepackage[caption=false]{subfig}
\newcommand{\ineq}[1]{\footnotesize$#1$\normalsize}{}
\usepackage[redeflists]{IEEEtrantools}
\begin{document}
\bstctlcite{IEEEexample:BSTcontrol}
\fancyhead{}
\title{A Design Methodology for Fault-Tolerant Computing using Astrocyte Neural Networks}

\author{Murat Isik}
\email{mci38@drexel.edu}
\authornote{Authors contributed equally to this research.}
\affiliation{%
  \institution{Drexel University}
  \city{Philadelphia}
  \state{PA}
  \country{USA}
  \postcode{19104}
}
\author{Ankita Paul}
\authornotemark[1]
\email{ap3737@drexel.edu}
\affiliation{%
  \institution{Drexel University}
  \city{Philadelphia}
  \state{PA}
  \country{USA}
  \postcode{19104}
}

\author{M. Lakshmi Varshika}
\authornotemark[1]
\email{lm3486@drexel.edu}
\affiliation{%
  \institution{Drexel University}
  \city{Philadelphia}
  \state{PA}
  \country{USA}
  \postcode{19104}
}

\author{Anup Das}
\email{anup.das@drexel.edu}
\affiliation{%
  \institution{Drexel University}
  \city{Philadelphia}
  \state{PA}
  \country{USA}
  \postcode{19104}
}
\begin{abstract}
We propose a design methodology to facilitate fault tolerance of deep learning models.
First, we implement
a many-core fault-tolerant \nm{} hardware design, where neuron and synapse circuitries in each \nm{} core are enclosed with \astro{} \ckts{}, the star-shaped glial cells of the brain that facilitate self-repair by restoring the spike firing frequency of a failed neuron
using a closed-loop retrograde feedback signal.
Next, we introduce \astro{s} in a deep learning model to achieve the required degree of tolerance to hardware faults.
Finally, we use a system software to partition the \astro{}-enabled model into clusters and implement them on the proposed \ftn{} \nm{} design. We evaluate this design methodology using seven deep learning inference models and show that it is both area- and power-efficient.
\end{abstract}



\maketitle

\section{Introduction}

Modern embedded systems are embracing \nm{} devices to implement spiking-based deep learning inference applications~\cite{cao2015spiking}.
A \nm{} device is designed as a many-core hardware, where each core consists of silicon \ckts{} to implement neurons and synapses~\cite{sentryos}.
Although technology scaling has provided a steady increase of performance, increased power densities (hence temperatures) and other scaling effects create an adverse impact on the reliability by increasing the likelihood of transient, intermittent, and permanent faults in the neuron and synapse \ckts{}~\cite{song2020case,reneu}. 
Hardware faults introduce errors in a trained deep learning model implemented on those \ckts{}, compromising 
inference quality (assessed using the accuracy metric).
Therefore, 
providing fault tolerance is 
a critical requirement for \nm{} devices.

Recent efforts to this end include software solutions such as model replication~\cite{ponzina2021e2cnns} and error prediction coding~\cite{park2020low}, and hardware solutions such as approximation~\cite{siddique2021exploring} and redundant mapping~\cite{yuan2021improving}.
For FPGA-based neuromorphic designs, fault tolerance can also be addressed using periodic scrubbing~\cite{santos2014criticality,venkataraman2014bit}.
In this work, we 
propose a complimentary approach to fault tolerance.
We exploit
the self-repair capability of the brain, which copes with damaged neurons using \astro{s}, the star-shaped glial cells 
of the brain~\cite{parpura1994glutamate}. 
Astrocytes generate an indirect retrograde feedback signal, which helps to restore the spike firing frequency of a failed neuron~\cite{nadkarni2007modeling}.



We propose a design methodology for \ftn{} neuromorphic computing, which consists of the following three components.

\begin{itemize}
    \item We propose a many-core \nm{} design where neurons in each core are enclosed with \astro{s} to facilitate self-repair of errors caused by logic and memory faults. 
    \item We introduce \astro{s} in a deep learning model to achieve a desired degree of tolerance to hardware faults.
    \item We propose a system software to partition an \astro{}-enabled inference model into clusters and implement them on the proposed \ftn{} \nm{} cores of the hardware. 
\end{itemize}

We evaluate our design methodology 
using seven deep learning inference models. 
Results show that the proposed design methodology is both area- and power-efficient, yet providing a high degrees of \ftc{} to randomly injected faults.



\section{Astrocyte Neural Networks}\label{sec:astrocyte}
\begin{figure}[h!]
	\centering
	\vspace{-10pt}
	\centerline{\includegraphics[width=0.89\columnwidth]{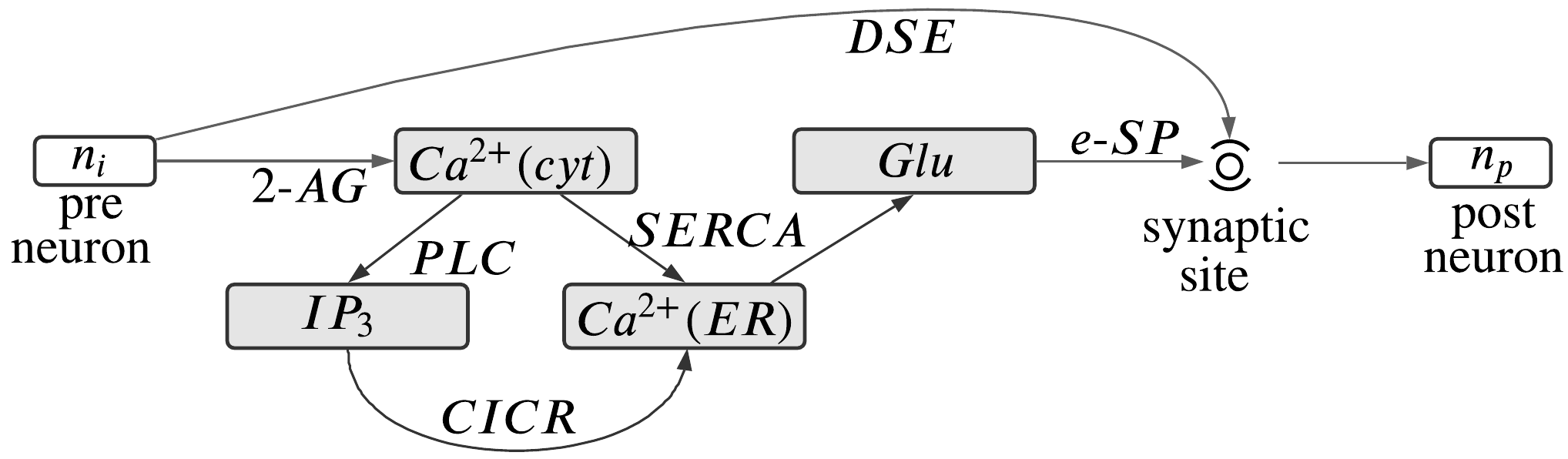}}
	\vspace{-10pt}
	\caption{Operation of an \astro{} (gray blocks).}
	\vspace{-10pt}
	\label{fig:astrocyte}
\end{figure}

Figure~\ref{fig:astrocyte} illustrates how an \astro{} regulates the neuronal activity at a synaptic site using a closed-loop feedback mechanism.

Astrocyte causes a transient increase of intracellular calcium (\ineq{Ca^{2+}}) levels, which serves as the catalyst for self-repair. \ineq{Ca^{2+}}-induced \ineq{Ca^{2+}} release (CICR) is the main mechanism to regulate \ineq{Ca^{2+}} in the healthy brain. CICR is triggered by inosital 1,4,5-triphosphate (\ineq{IP_3}), which is produced upon astrocyte activation. To describe the operation of the astrocyte, let \ineq{\delta(t-\tau)} be a spike at time \ineq{\tau} from the neuron \ineq{n_i}. This spike triggers the release of 2-arachidonyl glycerol (2-AG), a type of endocannabinoid responsible for stimulating the cytosolic calcium \ineq{Ca^{2+}} (cyt). The quantity of 2-AG produced is governed by the ordinary differential equation (ODE)
\begin{equation}
    \label{eq:2ag}
    \footnotesize \frac{dAG}{dt} = \frac{-AG}{\tau_{AG}} + r_{AG}\cdot\delta(t-\tau),
\end{equation}
where \ineq{AG} is the quantity of 2-AG, \ineq{\tau_{AG}} is the rate of decay and \ineq{r_{AG}} is the rate of production of 2-AG.

On one pathway, the cytosolic calcium is absorbed by the endoplasmic reticulum (ER) via the Sarco-Endoplasmic-Reticulum \ineq{Ca^{2+}}-ATPase (SERCA) pumps, and on the other pathway, the cytosolic calcium enhances the Phospholipase C (PLC) activation process. This event increases \ineq{IP_3} production and ER intracellular calcium release via the CICR mechanism.

The intracellular astrocytic calcium dynamics control the glutamate (Glu) release from the astrocyte, which is governed by
\begin{equation}
    \label{eq:glu}
    \footnotesize \frac{dGlu}{dt} = \frac{-Glu}{\tau_{Glu}} + r_{Glu}(t-t_{Ca}),
\end{equation}
where \ineq{\tau_{Glu}} is the rate of decay and \ineq{r_{Glu}} is the rate of production of glutamate, and \ineq{t_{Ca}} is time at which \ineq{Ca^{2+}} crosses the release threshold. The glutamate generates e-SP, the indirect signal to the synaptic site. e-SP is related to Glu using the following ODE
\begin{equation}
    \label{eq:esp}
    \footnotesize \frac{deSP}{dt} = \frac{-eSP}{\tau_{eSP}} + \frac{m_{eSP}}{\tau_{eSP}}Glu(t),
\end{equation}
where \ineq{\tau_{eSP}} is the decay rate of e-SP and \ineq{m_{eSP}} is a scaling factor.

Finally, there exists a direct signaling pathway (DSE) from neuron \ineq{n_i} to the synaptic site. The DSE is given by
\begin{equation}
    \label{eq:dse}
    \footnotesize DSE = -K_{AG}\cdot AG(t),
\end{equation}
where \ineq{K_{AG}} is a constant.
Overall, the synaptic transmission probability (PR) at the synaptic site is
\begin{equation}
    \label{eq:pr}
    \footnotesize PR(t) = PR(0) + PR(0)\left(\frac{DSE(t) + eSP(t)}{100}\right)
\end{equation}

\begin{figure}[h!]%
    \centering
    \vspace{-10pt}
    \subfloat[Original network.\label{fig:original_connection}]{{\includegraphics[width=3.2cm]{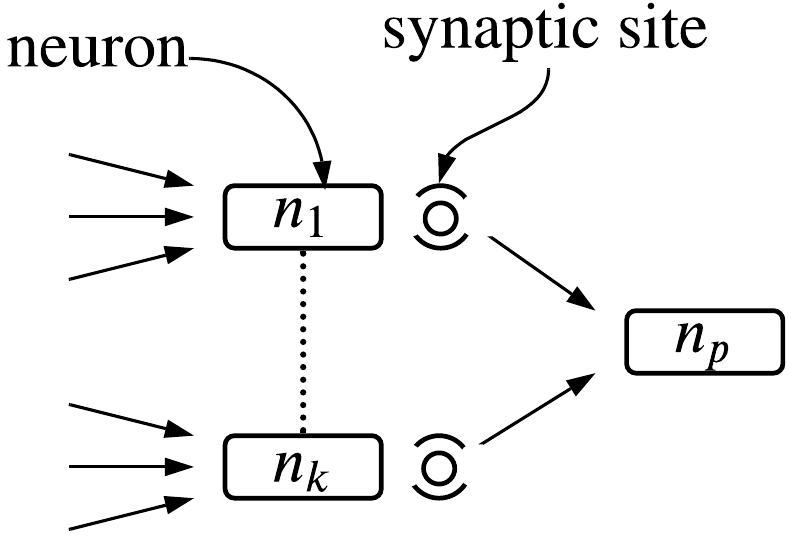} }}%
    \hfill
    \subfloat[Astrocyte-modulated network.\label{fig:astrocyte_modulation}]{{\includegraphics[width=4.9cm]{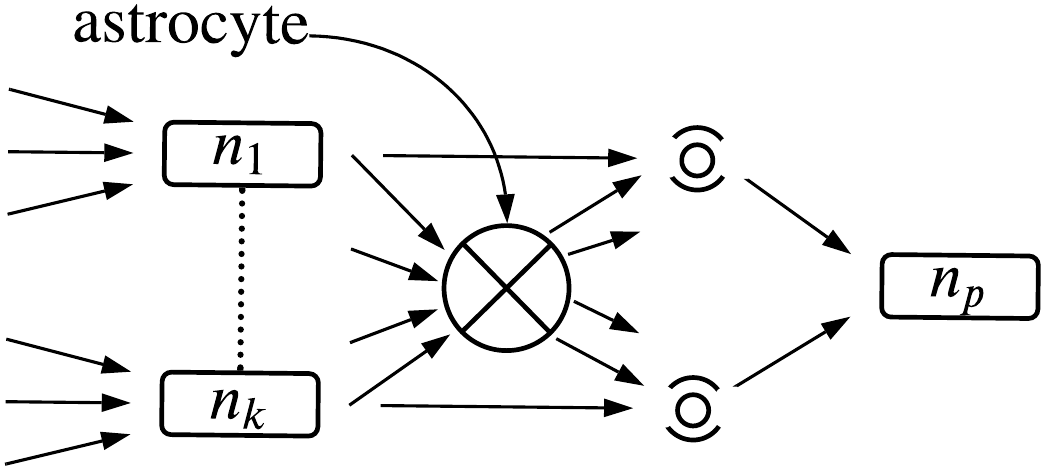} }}%
    \vspace{-10pt}
    \caption{Inserting an \astro{} in a neural network.}%
    \label{fig:astrocyte_neural_network}%
    \vspace{-10pt}
\end{figure}

In the brain, each \astro{} encloses multiple synapses connected to a neuron. Figure~\ref{fig:original_connection} shows an original network of neurons, while Figure~\ref{fig:astrocyte_modulation} shows these neurons enclosed using an \astro{}.

To understand the self-repair mechanism, 
consider neuron \ineq{n_i} in Fig.~\ref{fig:astrocyte} fails to fire a spike. 
Without the \astro{}, the spike firing rate at the synaptic site would decreases. However, because of the \astro{}, 2-AG production reduces (Eq.~\ref{eq:2ag}). This increases the DSE (Eq.~\ref{eq:dse}). Therefore, the PR increases (Eq.~\ref{eq:pr}) along with an increase of the spike firing frequency at the synaptic site.

\begin{figure}[h!]
	\centering
	\vspace{-10pt}
	\centerline{\includegraphics[width=0.99\columnwidth]{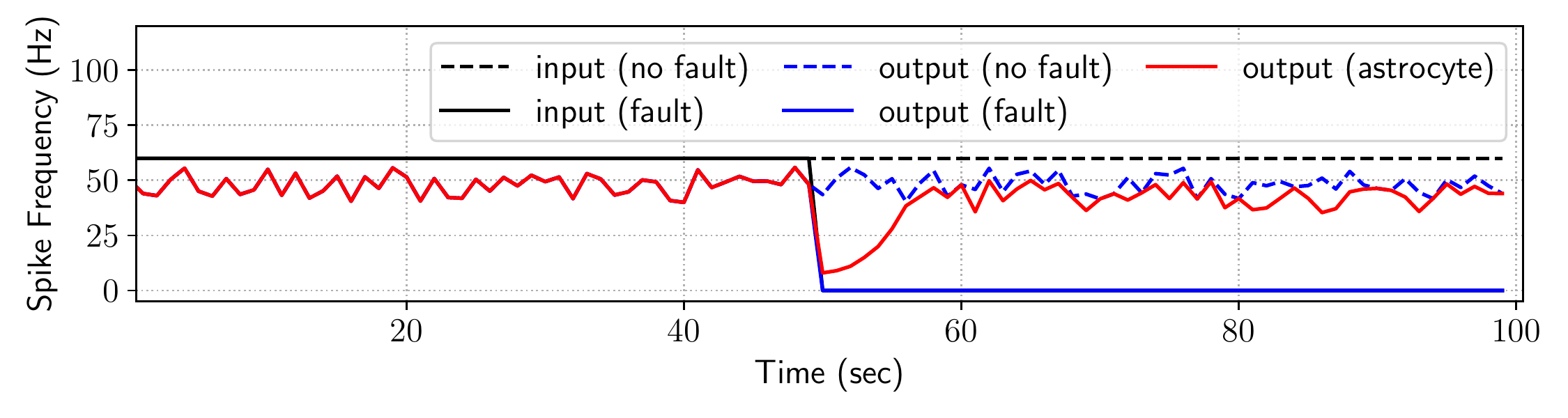}}
	\vspace{-10pt}
	\caption{Self-repair mechanism of an \astro{}.}
	\vspace{-10pt}
	\label{fig:error_recovery}
\end{figure}

Figure~\ref{fig:error_recovery} illustrates the self-repair mechanism. The input neuron \ineq{n_i} is excited with Poisson spike events having a mean spike rate of 60Hz. We interrupt the input at around 50 sec. We observe that the firing frequency at the synaptic site connected to \ineq{n_i} drops to 0. This is indicated with the label \textit{output (fault)}. Using \astro{}, the firing frequency can be restored partially as illustrated using the label \textit{output (\astro{})}.


\section{Proposed Design Methodology}\label{sec:design_methodology}
\subsection{Novel Hardware With Astrocyte Circuitries}
\begin{figure}[h!]
	\centering
	\vspace{-10pt}
	\centerline{\includegraphics[width=0.99\columnwidth]{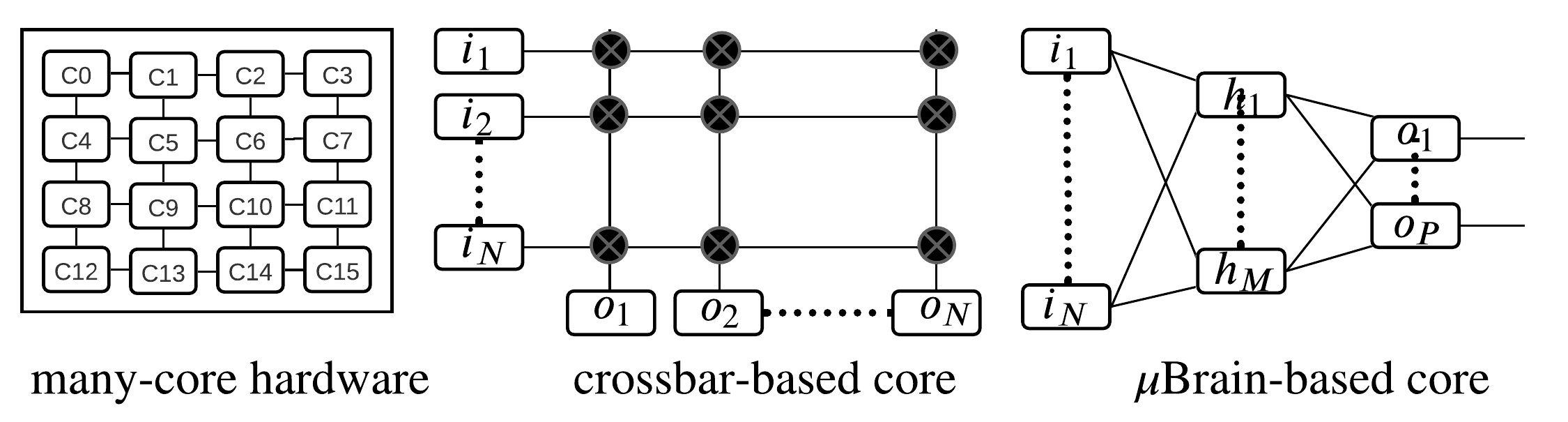}}
	\vspace{-10pt}
	\caption{Baseline architecture of a neuromorphic hardware.}
	\vspace{-10pt}
	\label{fig:architecture}
\end{figure}

Figure~\ref{fig:architecture} shows the architecture of a many-core neuromorphic hardware (left sub-figure). We take the example of two recent designs -- DYNAPs~\cite{dynapse}, where each core consists of an \textit{N}\ineq{\times}\textit{N} crossbar with \ineq{N} pre-synaptic neurons connected to \ineq{N} post-synaptic neurons (middle sub-figure), and \mubrain{}~\cite{sentryos}, where each core consists of neurons that are organized in three layers with \ineq{N} neurons in layer 1, \ineq{M} neurons in layer 2, and \ineq{P} neurons in layer 3 (right sub-figure).

\begin{figure}[h!]
	\centering
	\vspace{-10pt}
	\centerline{\includegraphics[width=0.99\columnwidth]{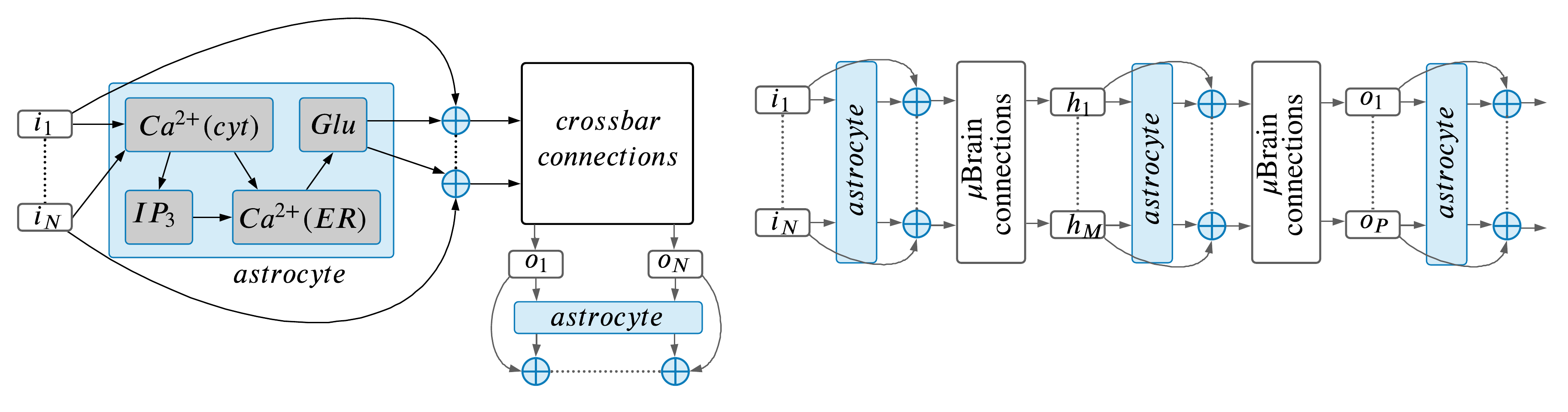}}
	\vspace{-10pt}
	\caption{Proposed design of crossbar (left) and \mubrain{} (right). }
	\vspace{-10pt}
	\label{fig:design_change}
\end{figure}

Figure~\ref{fig:design_change} illustrates our proposed changes to a baseline crossbar (left sub-figure) and a baseline \mubrain{} (right sub-figure) design.

\subsection{Software Mapping Framework}\label{sec:software}
A single \nm{} core can implement only a limited number of neurons and synapses. A \ineq{128\times128} crossbar core consists of 128 input and 128 output neurons, while a \mubrain{} core consists of 256 neurons in layer 1, 64 neurons in layer 2, and 16 neurons in layer 3.
We use a distance-based heuristic~\cite{sentryos} to partition an inference model into clusters, where each cluster can be implemented on a core of the hardware.\footnote{Apart from distance-based heuristic, recently heuristic graph partitioning approaches are also proposed in literature~\cite{spinemap,dfsynthesizer_pp,huynh2022implementing}.}
It sorts all neurons of a model based on their distances from output neurons. For \mubrain{} (crossbar) mapping, it groups all neurons with distance less than or equal to 2 (1) into clusters considering the resource constraint of a core. In the next iteration, it removes already clustered neurons from the model, recalculates neuron distances, and groups remaining neurons to generate the next set of clusters. The process is repeated until all neurons are clustered. By incorporating hardware constraints, we ensure that a cluster can fit onto the target core architecture.


\subsection{Astrocyte-Enabled Inference Model}

We introduce the following notations.
\begin{footnotesize}
\begin{align*}
G_M(C,E) =&~\text{Inference model with } C \text{ clusters and } E \text{ edges}\\
G_A(C_A,E) =&~\text{Astrocyte-enabled model with } C_A \text{ clusters and } E \text{ edges}\\
L =&~\text{Layers of a core. } L = \{L_x,L_y\} \text{ (crossbar) and } L = \{L_x,L_y,L_z\} (\mu\text{Brain})
\end{align*}
\end{footnotesize}
\normalsize

Algorithm~\ref{alg:astrocyte} shows the pseudo-code to insert \astro{s} in clusters of an inference model \ineq{G_M}. First, it organizes the neurons of a cluster into two (for crossbar) or three (for \mubrain{}) layers (line 2). 
Next, for each layer it uses the \texttt{ARES} framework~\cite{reagen2018ares} to insert \ineq{N_r} random errors, one at a time and record the corresponding accuracy (line 5). 
If the minimum accuracy \ineq{a_{min}} is lower than a threshold \ineq{a_{th}}, it adds an \astro{} to the layer (lines 6-8). Otherwise, it exits and analyzes the next layer (lines 8-9). In allocating \astro{s} to a layer, if more than one \astro{s} are needed, then its distributes neurons of the layer equally amongst the \astro{s}. \ineq{N_r} and \ineq{a_{th}} are user defined parameters and they are empirically set to 10,000 and \ineq{a_{o}}, respectively, where \ineq{a_o} is the baseline accuracy of the model without error. Finally, the \astro{}-enabled model (\ineq{G_A}) is returned.

\vspace{-10pt}
\begin{algorithm}[h]
	\scriptsize{
 		\KwIn{\ineq{G_{M}= (\textbf{C},\textbf{E})}}
 		\KwOut{\ineq{G_{A}= (\textbf{C}_A,\textbf{E})}}
 		\For(\tcc*[f]{For each cluster in $C$}){$C_k\in C$}{
 		    $C_k = \{C_k^x,C_k^y,C_k^z\}$\tcc*[r]{arrange neurons \& synapses of $C_k$ into three layers for $\mu$Brain core. For crossbar mapping, $C_k = \{C_k^x,C_k^y\}$.}
 		     \For(\tcc*[f]{For each layer in $C_k$}){$C_k^i\in C_k$}{
 		        \While(\tcc*[f]{Run until all neurons of the layer are protected against randomly injected errors}){(true)}{
 		            Insert $N_r$ random errors using \texttt{ARES} and evaluate the minimum accuracy $a_{min}$\;
 		            \uIf(\tcc*[f]{Min accuracy is less than threshold.}){$a_\text{min}<a_{th}$}{
 		                $C_k^i = C_k^i\cup$ \texttt{A}\tcc*[r]{Add an \astro{}.}
 		            }
 		            \Else{
 		                exit\;
 		            }
 		        }
 		     }
 		}
 	}
	\caption{Inserting \astro{s} in clusters of a model.}
	\label{alg:astrocyte}
\end{algorithm}
\vspace{-20pt}

\section{Evaluation}\label{sec:evaluation}
Our simulation framework consists of the following.
\begin{itemize}
    \item \texttt{QKeras:} to train 2-bit quantized deep learning models.
    \item \texttt{PyCARL\cite{pycarl}:} to generate spiking inference models.
    \item \texttt{Brian 2~\cite{stimberg2019modeling}:} for \astro{} modeling.
    \item \texttt{ARES~\cite{reagen2018ares}:} for fault simulations.
    \item \texttt{Xilinx Vivado:} for FPGA synthesis.
\end{itemize}

\subsection{Astrocyte Area and Power}
\begin{figure}[h!]
	\centering
	\vspace{-10pt}
	\centerline{\includegraphics[width=0.80\columnwidth]{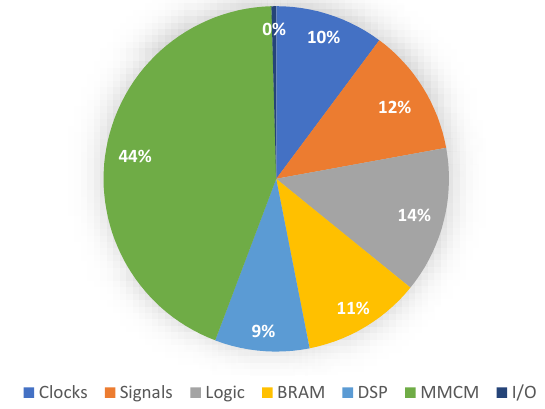}}
	\vspace{-10pt}
	\caption{Power consumption of astrocyte, distributed into clocks, signals,BRAMs,DSPs,MMCM, and I/Os.}
	\vspace{-15pt}
	\label{fig:astro_power}
\end{figure}

We implemented the \astro{} design, the baseline \mubrain{} and crossbar designs on Xilinx VCU128 development board (see Table~\ref{tab:area_astro}). 
We observe that although an \astro{} \ckt{} is smaller than the size of a \mubrain{} (336 neurons) and a crossbar (256 neurons), it is in fact, significantly larger and consumes significantly higher power than a single neuron \ckt{}. Furthermore, an \astro{} \ckt{} uses more flip flops (FF), slices, and lookup tables (LUTs) than the two baseline designs. The higher area of the two baseline designs are due to the use of more block RAMs (BRAMs). 
The power consumption of an \astro{} design is shown in Figure~\ref{fig:astro_power}, distributed into clocks, signals, logic, DSP, BRAM, MMCM, and I/O. 

\vspace{-10pt}
\begin{table}[h!]
	\renewcommand{\arraystretch}{0.8}
	\setlength{\tabcolsep}{2pt}
	\caption{Implementation of an \astro{} and the baseline \mubrain{}~\cite{sentryos} and crossbar~\cite{dynapse} designs on Xilinx VCU128.}
	\label{tab:area_astro}
	\centering
	\begin{threeparttable}
	{\fontsize{6}{10}\selectfont
	    \vspace{-10pt}
		\begin{tabular}{c|ccc}
		    \hline
		    & \textbf{\mubrain{}~\cite{sentryos}} & \textbf{Crossbar~\cite{dynapse}} & \textbf{Astrocyte}\\
		    \hline
		    Neurons & 336 & 256 & --\\
		    Synapses & 17,408 & 16,384 & --\\
		    \hline
		    Operating Frequency & 100MHz & 100MHz & 100MHz \\
		    BRAM & 48 & 32 & 4 \\
		    DSP & 0 & 0 & 4 \\
		    FF & 129 & 86 & 2,368 \\
		    Slice & 117 & 78 & 670 \\
		    LUT & 114 & 76 & 1,345 \\
		    \hline
		    FPGA Utilization & 49\% & 40\% & 12\%\\
		    \hline
		    Power & 4.64W & 4.53W & 0.538 W \\
			\hline
	    \end{tabular}}
	\end{threeparttable}
	\vspace{-10pt}
\end{table}

\subsection{Fault Tolerance}
\begin{figure}[h!]
	\centering
	\vspace{-10pt}
	\centerline{\includegraphics[width=0.99\columnwidth]{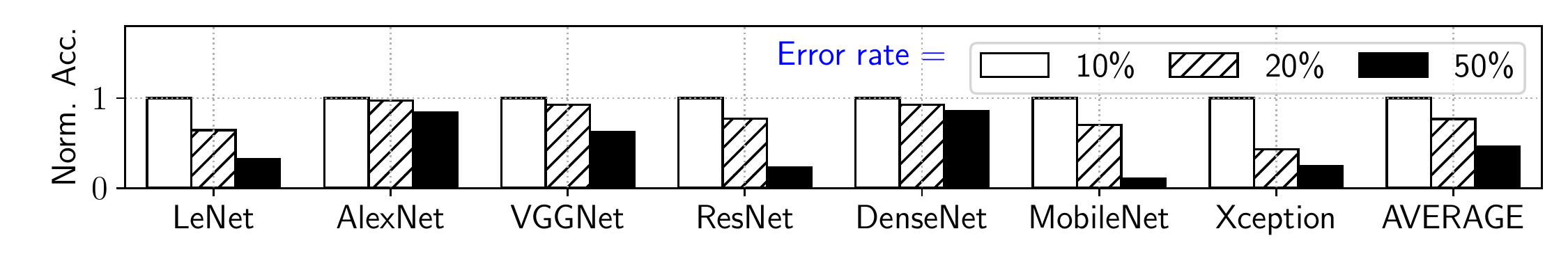}}
	\vspace{-10pt}
	\caption{Normalized accuracy for different error rates.}
	\vspace{-10pt}
	\label{fig:accuracy_final}
\end{figure}

Figure~\ref{fig:accuracy_final} plots the accuracy, normalized to the replication technique, of each evaluated model for 10\%, 20\%, and 50\% of parameters in error. These errors are injected randomly using the \texttt{ARES} framework~\cite{reagen2018ares} and the reported results are average of 10 runs.
With 10\% error rate, there are only a few errors per cluster. Therefore, most errors can be masked by \astro{s} that are inserted into each model cluster. So, we see no accuracy drop. With higher error rates, the accuracy is lower. This is because of the increase in parameter errors in each cluster. 
Errors in multiple neurons of an enclosed \astro{} impact its ability to restore the spike frequency, causing a significant amount of accuracy drop. On average, the accuracy is 23\% and 54\% lower for error rate of 20\% and 50\%, respectively.

\subsection{Design Tradeoffs}
\begin{figure}[h!]
	\centering
	\vspace{-10pt}
	\centerline{\includegraphics[width=0.99\columnwidth]{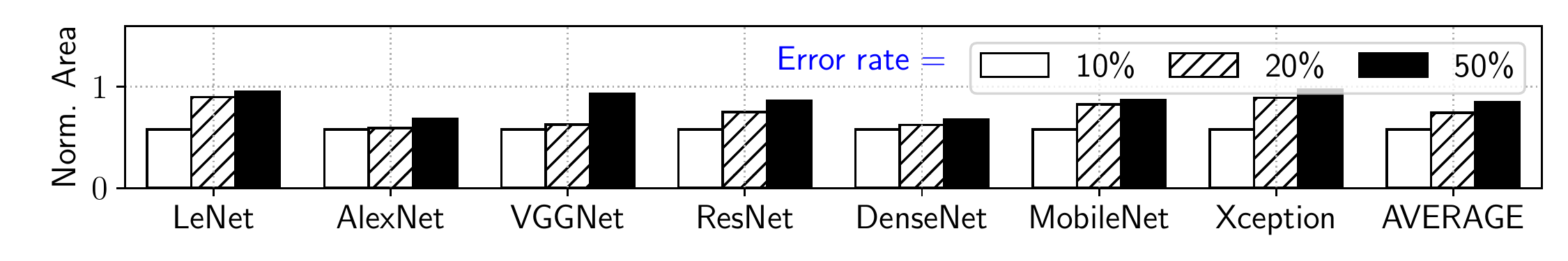}}
	\vspace{-10pt}
	\caption{Normalized area for different error rates.}
	\vspace{-10pt}
	\label{fig:area_scalability}
\end{figure}

Figure~\ref{fig:area_scalability} shows the area of a \mubrain{}-based design normalized to the replication technique for three error rates -- 10\%, 20\%, and 30\%. 
The accuracy constraint is set as the accuracy without error.
This accuracy constraint is achieved for 10\% error rate using our baseline design. So there is no area overhead. For 20\% and 50\% error rates, more \astro{s} are needed to achieve the accuracy constraint. On average, the proposed design requires 28\% and 49\% higher area for 20\% and 50\% error rate, respectively. 

\subsection{Model Area}
Table~\ref{tab:area} reports the design area for each of the evaluated deep learning inference models using 1) model replication technique~\cite{ponzina2021e2cnns}, 2) redundant mapping technique~\cite{yuan2021improving}, and 3) the proposed design methodology. Design areas are reported for both the \mubrain{}-based core~\cite{sentryos} and the crossbar-based core~\cite{dynapse}. All results are normalized to the \mubrain{}-based design implementing the LeNet model using the model replication technique. We make three key observations.

\vspace{-10pt}
\begin{table}[h!]
	\renewcommand{\arraystretch}{0.8}
	\setlength{\tabcolsep}{2pt}
	\caption{Design area compared to model replication~\cite{ponzina2021e2cnns} and redundant mapping~\cite{yuan2021improving}.}
	\label{tab:area}
	\vspace{-10pt}
	\centering
	\begin{threeparttable}
	{\fontsize{6}{10}\selectfont
		\begin{tabular}{c|cc|cc|cc}
		    \hline
		    & \multicolumn{2}{|c|}{\textbf{Model}} & \multicolumn{2}{|c|}{\textbf{Redundant}} & \multicolumn{2}{|c}{\textbf{Proposed}}\\
		    & \multicolumn{2}{|c|}{\textbf{Replication~\cite{ponzina2021e2cnns}}} & \multicolumn{2}{|c|}{\textbf{Mapping~\cite{yuan2021improving}}} & \multicolumn{2}{|c}{\textbf{Design}}\\
			\cline{2-7}
			& \textbf{\mubrain{}} & \textbf{crossbar} & \textbf{\mubrain{}} & \textbf{crossbar} & \textbf{\mubrain{}} & \textbf{crossbar}\\
			\hline
			LeNet & 1.0 & 0.8 & -- & 0.7 & 0.5 & 0.4 \\
			AlexNet & 79.0 & 68.5 & -- & 54.8 & 39.2 & 33.1\\
			VGGNet & 62.9 & 54.6 & -- & 43.7 & 31.2 & 26.4\\
			ResNet & 1.1 & 0.9 & -- & 0.8 & 0.6 & 0.5\\
			DenseNet & 13.5 & 11.7 & -- & 9.4 & 6.7 & 5.7\\
			MobileNet & 4.4 & 3.8 & -- & 3.0 & 2.2 & 1.8\\
			Xception & 40 & 34.7 & -- & 27.7 & 19.9 & 16.8\\
			\hline
	    \end{tabular}}
	\end{threeparttable}
\end{table}
\vspace{-10pt}

First, design area is larger for models with higher number of parameters. This is because models with more parameters require more clusters (cores), which increases the design area. 
Second, the redundant mapping technique is only applicable to crossbar-based designs.
Therefore, results for the \mubrain{}-based design are not provided.
Third, for the \mubrain{}-based design, the proposed design methodology results in 50\% lower area than the replication technique. For the crossbar-based design, it results in 51.6\% lower area than the replication technique and 39.5\% lower area than the redundant mapping technique. These improvements are because implementing a few \astro{s} in a baseline \mubrain{} and crossbar designs is area-efficient than 1) replicating model clusters, which requires more cores to implement a model, and 2) redundant mapping, which requires larger crossbars to implement each cluster.

\subsection{Model Power}
\begin{figure}[h!]
	\centering
	\centerline{\includegraphics[width=0.99\columnwidth]{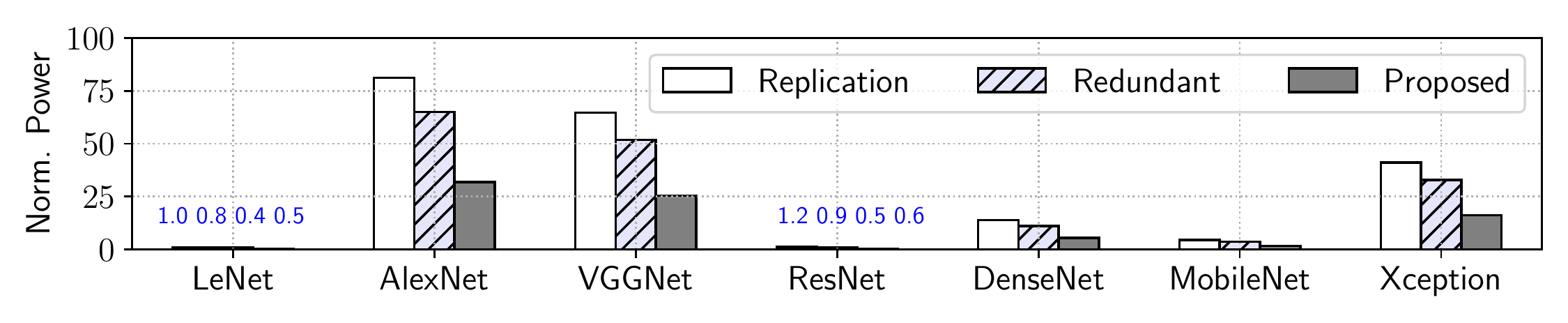}}
	\vspace{-15pt}
	\caption{Power consumption.}
	\vspace{-10pt}
	\label{fig:power}
\end{figure}

Figure~\ref{fig:power} reports the power for each evaluated model on a crossbar-based design using the three evaluated approaches. Power numbers for each core is calculated based on the static power of the design and the activation of the synaptic weights in the core~\cite{twisha_energy}. We make two key observations. First, power is higher for models such as AlexNet, VGGNet, and Xception due to higher number of model parameters. Second, on average, power using the proposed design methodology is 60\% lower than replication technique and 50\% lower than redundant mapping technique. For \mubrain{}-based design (not shown here for space limitations), power using the proposed methodology is 60\% lower than the replication technique.

\section{Conclusions}\label{sec:conclusions}
We propose a design methodology for \ftn{} \nm{} computing. First, we propose a novel design, where a core consists of neuron, synapse, and \astro{} \ckts{}. Each \astro{} encloses multiple neurons to facilitate self-repair of a failed neuron.
Next, we insert \astro{s} in an inference model to achieve the desired degree of fault tolerance.
Finally, we propose a system software framework to map \astro{}-enabled inference model to the proposed fault-tolerant many-core design.
We evaluate the proposed design methodology using several deep learning models on the \ftn{} implementation of two baseline \nm{} designs. 
We show that the proposed design methodology is both area and power-efficient, yet providing similar degrees of fault tolerance compared to existing approaches.
\begin{acks}
This work is supported by the National Science Foundation Faculty Early Career Development Award CCF-1942697 (CAREER: Facilitating Dependable Neuromorphic Computing: Vision, Architecture, and Impact on Programmability).
\end{acks}

\bibliographystyle{IEEEtranSN}
\bibliography{commands,disco,external}


\end{document}